\documentclass[sigconf]{acmart}

\pdfoutput=1

\usepackage{algorithm}
\usepackage{algorithmicx}
\usepackage{algpseudocode}
\usepackage{amsmath}
\usepackage{booktabs}
\usepackage{multirow}
\usepackage[normalem]{ulem}
\usepackage{graphicx}
\usepackage{subfigure}

\AtBeginDocument{%
  \providecommand\BibTeX{{%
    \normalfont B\kern-0.5em{\scshape i\kern-0.25em b}\kern-0.8em\TeX}}}

\copyrightyear{2023}
\acmYear{2023}
\setcopyright{acmlicensed}
\acmConference[MM '23] {Proceedings of the 31st ACM International Conference on Multimedia}{October 29--November 3, 2023}{Ottawa, ON, Canada.}
\acmBooktitle{Proceedings of the 31st ACM International Conference on Multimedia (MM '23), October 29--November 3, 2023, Ottawa, ON, Canada}
\acmPrice{15.00}
\acmISBN{979-8-4007-0108-5/23/10}
\acmDOI{10.1145/3581783.3612027}

\begin{document}

\title{Federated Deep Multi-View Clustering with Global Self-Supervision}

% \affiliation{%

\author{Xinyue Chen}
\affiliation{%
  \institution{School of Computer Science and Engineering, University of Electronic Science and Technology of China}
  \city{Chengdu}
  \country{China}
  }
\email{martinachen2580@gmail.com}

\author{Jie Xu}
\affiliation{%
  \institution{School of Computer Science and Engineering, University of Electronic Science and Technology of China}
  \city{Chengdu}
   \country{China}
  }
\email{jiexuwork@outlook.com}

\author{Yazhou Ren}
\authornote{Corresponding author.}
\affiliation{%
  \institution{School of Computer Science and Engineering, University of Electronic Science and Technology of China}
  \city{Chengdu}
  \country{China}
  }
\email{yazhou.ren@uestc.edu.cn}

\author{Xiaorong Pu}
\affiliation{%
  \institution{School of Computer Science and Engineering, University of Electronic Science and Technology of China}
  \city{Chengdu}
  \country{China}
  }
\email{puxiaor@uestc.edu.cn}

\author{Ce Zhu}
\affiliation{%
  \institution{School of Information and Communication Engineering, University of Electronic Science and Technology of China}
  \city{Chengdu}
  \country{China}
  }
\email{eczhu@uestc.edu.cn}

\author{Xiaofeng Zhu}
\affiliation{%
  \institution{School of Computer Science and Engineering, University of Electronic Science and Technology of China}
  \city{Chengdu}
  \country{China}
  }
\email{seanzhuxf@gmail.com}

% \author{Xiaofeng Zhu}
% \affiliation{%
%   \institution{University of Electronic Science and Technology of China}
%   \city{Chengdu}
% \email{seanzhuxf@gmail.com}

\author{Zhifeng Hao}
\affiliation{%
  \institution{College of Science, Shantou University}
  \city{Shantou}
  \country{China}
  }
\email{haozhifeng@stu.edu.cn}

\author{Lifang He}
\affiliation{%
  \institution{Department of Computer Science and Engineering, Lehigh University}
  \city{Bethlehem}
  % \state{PA}
  \country{USA}
  }
\email{lih319@lehigh.edu}

% \renewcommand{\shortauthors}{Chen and Xu, et al.}
\renewcommand{\shortauthors}{Xinyue Chen et al.}

\begin{abstract}
  Federated multi-view clustering has the potential to learn a global clustering model from data distributed across multiple devices. In this setting, label information is unknown and data privacy must be preserved, leading to two major challenges. First, views on different clients often have feature heterogeneity, and mining their complementary cluster information is not trivial. Second, the storage and usage of data from multiple clients in a distributed environment can lead to incompleteness of multi-view data. To address these challenges, we propose a novel federated deep multi-view clustering method that can mine complementary cluster structures from multiple clients, while dealing with data incompleteness and privacy concerns. 
  Specifically, in the server environment, we propose sample alignment and data extension techniques to explore the complementary cluster structures of multiple views. 
  % We use global prototypes based on sample commonality and view versatility to impute incomplete data. 
  The server then distributes global prototypes and global pseudo-labels to each client as global self-supervised information. In the client environment, multiple clients use the global self-supervised information and deep autoencoders to learn view-specific cluster assignments and embedded features, which are then uploaded to the server for refining the global self-supervised information.
  Finally, the results of our extensive experiments demonstrate that our proposed method exhibits superior performance in addressing the challenges of incomplete multi-view data in distributed environments.
  % Finally, extensive experiments indicate that our method achieves superior performance on incomplete multi-view data in distributed environments.
  
\end{abstract}





\maketitle

\section{Introduction}

\begin{figure}[!h]
    \centering
    \includegraphics[width=8cm]{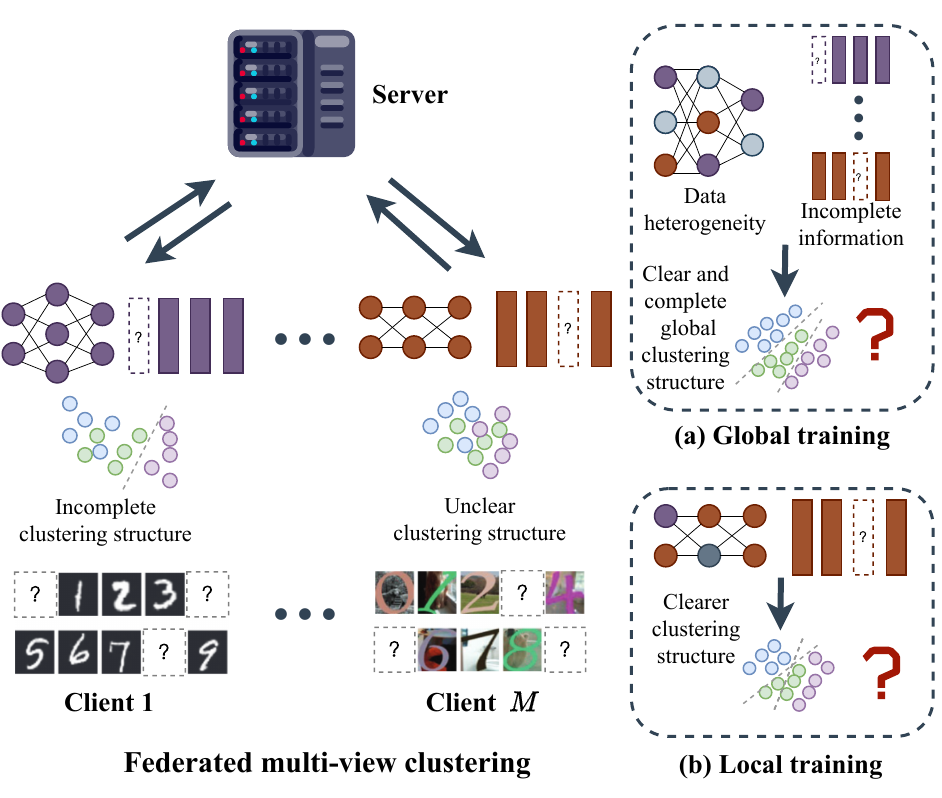}
    \caption{Problem illustration of federated multi-view clustering. (a) During global training, how can the server address feature heterogeneity and incomplete information in multi-view data to obtain a clear global clustering structure?
    % In global training, server how to face the feature heterogeneity and incomplete information of multi-view data and obtain a clear global clustering structure?
    (b) During local training, how can we alleviate the unclear clustering structure of the client with low-quality views?}
    % where the low quality views are located?}
    \label{fig:problem}
\end{figure}
Multimedia technologies have led to the emergence of a large amount of multi-view or multi-modal data, which often lack label information \cite{bickel2004multi,CCGC,GCC-LDA,xihong_ICL_SSL}. To explore useful consistent and complementary information among multiple views in an unsupervised manner, researchers have proposed various multi-view clustering methods \cite{li2021incomplete,ren2022diversified,xu2023untie}. However, these methods typically operate in a centralized environment and cannot handle isolated data stored in various distributed devices/silos due to privacy concerns in industry competition. Fortunately, federated learning offers a potential solution for such scenarios by enabling the training of a unified model without exposing sensitive data stored on individual devices.

% Multi-view clustering (MVC) \cite{bickel2004multi} has attracted wide attention due to its ability to leverage the consistent and complementary information among multiple views to improve clustering performance. Some MVC methods use different strategies, such as subspace clustering-based \cite{li2019reciprocal,zheng2020feature,ren2022diversified} or graph-based \cite{tang2020cgd,li2021incomplete,ma2022simultaneous}, and focus on exploring the shared representation of multiple views. Typically, these methods operate in centralized environment. 
% However, due to the growing storage and computing power of distributed devices, industry competition, privacy security, etc., sometimes it is impossible to collect and use the data in various devices or silos. 
% Federated learning provides a possible solution when isolated data scattered across devices or silos cannot be centrally trained to produce a unified model. \cite{konevcny2016federated,mcmahan2017communication,huang2022learn}.

% To this end, federated multi-view learning \cite{huang2022efficient,che2022federated} is an emerging machine learning paradigm and has attracted increasing attention in recent years.
% Federated multi-view learning aims to learn a global model from multiple views’ data distributed across different devices, which usually are combined with previous machine learning methods, such as multi-view matrix factorization \cite{flanagan2021federated,huang2022efficient}, ensemble learning \cite{che2022federated,feng2020multi}, and deep models \cite{huang2020federated,xu2021fedmood}.

Federated multi-view learning \cite{liu2020systematic,zhu2021federated} is a relatively new machine learning paradigm that has gained significant attention in recent years. It is designed to learn a global model from multi-view data that are distributed across different devices, and it often incorporates existing machine learning methods, such as multi-view matrix factorization \cite{flanagan2021federated,huang2022efficient}, ensemble learning \cite{che2022federated,feng2020multi}, and deep models \cite{huang2020federated,xu2021fedmood}. By combining these methods with the federated learning approach, it becomes possible to address the challenges posed by distributed multi-view data, such as data privacy and feature heterogeneity.

% (1) Some FedMVL methods \cite{flanagan2021federated,huang2022efficient} extend the federated learning framework to multi-view matrix factorization, and aggregate or select the low-rank matrix for each view on the server.
% (2) Ensemble-based FedMVL \cite{che2022federated}, which usually trains a single learner locally with data from each client and exploits the differences among multiple learners on the server to improve the learning performance.
% (3) Deep learning techniques are applied to FedMVL, including but not limited to (a) Deep structured semantic models (DSSM) is used to map users and items to a shared semantic space within a federated multi view setting \cite{huang2020federated}; (b) DeepMood architecture is used for late fusion in a federated learning setting at the session level\cite{xu2021fedmood}.

% Federated multi-view clustering has the potential to learn a global clustering model from multiple views' data distributed across different devices. As label information is unknown and data privacy need to be preserved, two challenges exist in this setting:
% 1) Views on different clients usually have statistical heterogeneity, and mining their complementary cluster information is not trivial.
% 2) In distributed environments, the data storage and usage of multiple clients might cause the incompleteness of multi-view data.

% As an import data processing method, clustering analysis with federated multi-view learning can handle multi-view/modal data without label information in distributed environments, but there is little research on this topic in the literature.
Clustering analysis with federated multi-view learning, known as federated multi-view clustering (FedMVC), has recently been shown to be an effective method for handling multi-view/modal data without label information in distributed environments. However, despite its potential, FedMVC is still a relatively underexplored area of research.
Addressing the challenges of FedMVC is crucial, and there are two main obstacles to overcome. 
Firstly, due to the feature heterogeneity of multi-view data and the complexity of clustering, traditional federated learning solutions struggle to identify complementary cluster structures. Even if each client can cluster local data separately, the feature heterogeneity of the datasets may obscure certain clusters that only become apparent when the data are combined. 
Some federated clustering methods \cite{dennis2021heterogeneity,stallmann2022towards} extend traditional clustering algorithms to federated learning settings, but they have limited capability to learn feature representations and struggle to handle complex heterogeneous multi-view data. 
Secondly, the data storage and usage of multiple clients in distributed environments can lead to incomplete multi-view data \cite{yang2022robust,balelli2022differentially}. 
For example, medical tests distributed across different healthcare institutions can be considered as different views, but patients do not undergo all the corresponding tests at each institution, which makes most methods based on information completeness assumptions unavailable in such scenarios.

\begin{figure*}[!t]
    \centering
    \includegraphics[width=16cm]{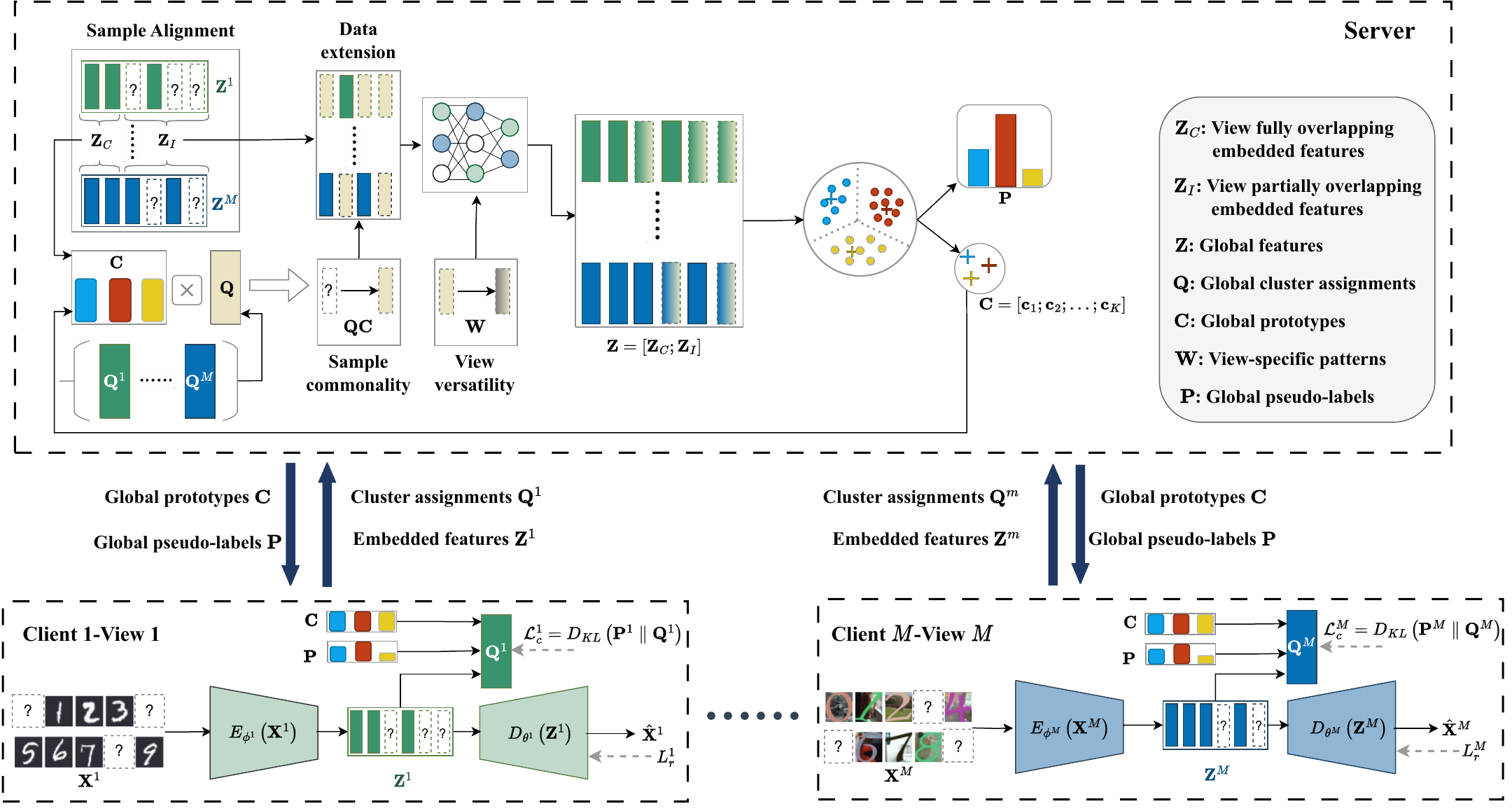}
    \caption{The framework of FedDMVC. It contains a server and $M$ clients.
    % It contains two main parts, $M$ clients and a server, where the multi-view data are distributed among $M$ clients, with each client owning a single view.
    % where there are $M$ views with multi-view data distributed among $M$ clients. 
    (1) Server: the server aggregates information uploaded by the clients and proposes sample alignment and data extension.
    % where incomplete data are imputed by using global prototypes $\mathbf{C}$ based on sample commonality and view versatility. 
    After that, the server proceeds to construct global features $\mathbf{Z}$, obtain global pseudo-labels $\mathbf{P}$, and explore the complementary cluster structures among multiple views.
    (2) Clients: For client $m$, we utilize the global self-supervised information and deep autoencoders to learn view-specific cluster assignments $\mathbf{Q}^m$ and embedded features $\mathbf{Z}^m$, which are then uploaded to the server for refining the global self-supervised information.}
    % (2) Clients: 
    % % each client contains an autoencoder and a clustering mapping. 
    % For client $m$, we optimize the reconstruction loss $\mathcal{L}_{\text {r }}^m$ and the clustering loss  $\mathcal{L} _{c}^{m}$ using global self-supervised information to obtain the embedding features $\mathbf{Z}^m$ and the view-specific clustering assignments $\mathbf{Q}^,$, which are then uploaded to the server for refining the global self-supervised information.}
    % (2) Server: the server aggregates the information uploaded by the clients, constructs global prototypes $\mathbf{C}$ and global cluster assignments $\mathbf{Q}$, discovers the global pseudo-labels $\mathbf{P}$ utilizing the global features $\mathbf{Z}$ through sample alignment and prototype-based imputation.}
    \label{fig:framework}
\end{figure*}

% To address the aforementioned issues, we propose a novel federated deep multi-view clustering method (FedDMVC), which is considered in the following scenario: a dataset with $M$ views is distributed across $M$ clients, and the samples constituting each client do not exactly overlap.
% % given a dataset with $M$ views, it is required to train machine learning models while $M$ views are distributed across $M$ devices or silos, i.e., 
% % Our proposed method primarily addresses the heterogeneity of features across multiple views and the issue of incomplete sample overlap among clients. 
% Our proposed method aims to mine complementary cluster structures from multiple clients by discovering and utilizing global self-supervised information within a federated learning setting. 
% The general framework is shown in Figure \ref{fig:framework}. 
% % There are $M$ clients and a server. 
% Specifically, the server discovers and distributes the global self-supervised information to each client, such as global prototypes and global pseudo-labels. Thereafter, multiple clients utilize the global self-supervised information and deep autoencoders to learn view-specific cluster assignments and embedded features, which are then uploaded to the server for the next iteration.
To overcome the challenges outlined above, we introduce a novel federated deep multi-view clustering method, FedDMVC. 
Our method is designed for scenarios where a dataset with $M$ views is distributed across $M$ clients, and the samples of each client do not have exact overlaps. 
The primary aim of our approach is to leverage global self-supervised information within a federated learning setting to extract complementary cluster structures from the data distributed across multiple clients. 
The general framework of FedDMVC is illustrated in Figure \ref{fig:framework}. 
Initially, the server distributes global self-supervised information, such as global prototypes and global pseudo-labels, to each client. 
Then, each client utilizes deep autoencoders and the global self-supervised information to learn view-specific cluster assignments and embedded features, which are then uploaded to the server for the next iteration.

In general, existing vertical federated learning methods involve parties sharing embeddings in a private manner \cite{bonawitz2016practical,castiglia2022compressed}, followed by a server model that captures the complex interactions of the embeddings.
% followed by a server model capturing the complex interactions of the embeddings.
For the proposed FedDMVC method, we additionally upload the cluster assignments of each client to the server. To address challenge 1, we construct global self-supervised information on the server to mitigate the heterogeneity of local datasets, and explore complementary cluster structures from multiple views across multiple clients.
To tackle challenge 2, we propose sample alignment and data extension techniques that leverage global prototypes and view-specific patterns to impute incomplete data based on sample commonality and view versatility.
% At this point, the server flows and integrates information from clients without exposing the original data, thereby discovering the global clustering structure. 
% In addition, the server distributes the constructed global self-supervised information to assist local training of each client, alleviating the heterogeneity of local datasets, thereby addressing the first challenge. To address the second challenge, we use the global prototypes $\mathbf{C}$ and global cluster assignments $\mathbf{Q}$ to effectively impute the missing samples.

% Different from existing methods, our contributions include the following three aspects:
% Our contributions differ from existing methods in three main aspects.
We summarize our contributions in this paper as follows:
\begin{itemize}
    % \vspace{-\baselineskip}
    \item We propose a novel federated deep multi-view clustering method that can effectively mine complementary cluster structures from multi-view data across multiple clients.
     % \item We design a comprehensive framework for federated multi-view clustering. This framework is able to adapt to the unsupervised multi-view environment and address the issue of incompleteness sample views. % c 1
    % \item Our method addresses the incompleteness of multi-view data across multiple clients by effectively expanding data from global prototypes and view-specific patterns based on sample commonality and view versatility.
    % \item We propose to expand data from global prototypes and view-specific patterns based on sample commonality and view versatility, to address the incompleteness of multi-view data in distributed environments.
    \item We propose a method to expand data from global prototypes and view-specific patterns based on sample commonality and view versatility, thereby addressing the incompleteness of multi-view data in distributed environments.
    % Our method expands the data effectively from global prototypes and view-specific patterns based on sample commonality and view versatility.
    % And use the extended data to mine global self-supervised information on the server.
    % from sample commonality and view versatility, while using the server to construct global self-supervised information without exposing the raw data of each client, and distributing enriched training labels to each client to improve the quality of each local model and obtain a high-quality global clustering structure. % c 2
    % \item Our method requires only two communications, avoiding the common issue of high communication cost in federated learning, saving computational resources and time. 
    \item Our proposed method can facilitate the flow and sharing of information among clients while ensuring privacy.
    % We enable the flow and sharing of information across clients in a privacy manner by sharing global self-supervised information. 
    % Extensive experiments on public datasets show that our method achieves superior performance over SOTAs. % others
    Extensive experiments conducted on public datasets demonstrate that our method outperforms state-of-the-art techniques.
\end{itemize}

\section{Related Work}

\subsection{Multi-View Clustering}

Multi-view clustering (MVC) methods aim to improve clustering performance by leveraging consistent and complementary information among multiple views. Traditional MVC methods utilize classical machine learning techniques such as non-negative matrix factorization (NMF), subspace and graph learning. 
Liu et al. \cite{liu2013multi} proposed an NMF-based method to handle multi-view data. Zhao et al. \cite{zhao2017multi} utilized a deep semi-NMF structure to extract more consistent information.
Similarly, subspace-based MVC methods achieve data clustering by exploring shared representations among multiple views.
% and similarity metric matrices among multiple views. 
For example, Li et al. \cite{li2019reciprocal} constructed a mutual multilayer subspace representation associated with latent representation to better recognize clustering structures. Zheng et al. \cite{zheng2020feature} introduced an effective feature cascaded multi-view subspace clustering to explore the consistency information of multi-view data.  
Graph-based MVC methods can exploit graph structure information to improve the recognition of clustering patterns. For instance, Wang et al. \cite{wang2016iterative} used multi-graph Laplacian regularized low-rank representation for multi-view graph clustering. Fan et al. \cite{fan2020one2multi} integrated self-supervised training with graph autoencoder reconstruction in a unified framework for attribute multi-view graph clustering.

In recent years, deep learning based MVC methods have been attracting increasing attention, among which MVC methods based on deep autoencoders have achieved remarkable achievements \cite{chen2020multi}. Deep autoencoders learn embedded features by optimizing the reconstruction loss between input and output \cite{xie2016unsupervised,peng2018structured}. They are usually combined with existing clustering methods to explore the unified clustering structure among multiple views. 
For example, Abavisani et al. \cite{abavisani2018deep} first used autoencoder architecture for multi-view subspace clustering. 
% Li et al. \cite{li2019deep} introduced adversarial training and combined it with an autoencoder, revealing the non-linear characteristics of multi-view data.
Although Xu et al. \cite{xu2022deep,xu2023adaptive} proposed deep imputation-free frameworks for addressing the incompleteness of multi-view data, data privacy issues in federated environments and the utilization of complementary information in incomplete parts of data have not been well studied.
% Xu et al. \cite{xu2021multi} used variational autoencoders (VAE) and introduced view-common and view-peculiar variables to learn separated visual representations.

Most traditional and deep MVC methods usually operate in centralized environments \cite{xu2022self}, which is difficult to handle data privacy leakage and data isolation issues. 
Although some distributed MVC methods \cite{cai2013multi,huang2018robust} have been proposed and can be applied in distributed environments, they are not suitable for addressing the unique challenges introduced by federated learning, such as feature heterogeneity and incompleteness of multi-view data across multiple clients. 
In this paper, we propose a novel federated deep multi-view clustering method that can solve the above issues.

% \vspace{0.2cm}
\subsection{Federated Multi-View Learning}

Federated multi-view learning presents effective solutions to the challenge of multi-view learning in federated environments. It can be roughly classified into three categories:
(1) Some FedMVL methods \cite{flanagan2021federated,huang2022efficient} extend the federated learning framework to include multi-view matrix factorization, which involves aggregating or selecting the low-rank matrix for each view on the server. 
For example, Flanagan et al. \cite{flanagan2021federated} combined multi-view matrix factorization with a federated learning framework used for personalized recommendations.
% and introduced a solution to the cold start problem. 
Huang et al. \cite{huang2022efficient} 
first considered the issues of high communication costs and proposed an NMF-based federated learning framework for the multi-view clustering task.
% proposed an NMF-based federated learning framework that first considered the issues of high communication costs and stragglers in federated multi-view learning. The method can also be applied for multi-view clustering task.
(2) Ensemble-based FedMVL methods, which usually train a single learner locally with data from each client and exploit the differences among multiple learners on the server to improve the learning performance. For instance, 
Feng et al. \cite{feng2020multi} proposed a multi-participant multi-class vertical federated learning framework that trains separate models for each participant.
% Feng et al. \cite{feng2020multi} extended the idea of multi-view learning and proposed the multi-participant multi-class vertical federated learning framework, which shares labels among participants in a privacy-preserving manner. 
Che et al. \cite{che2022federated} proposed a generic federated multi-view learning framework that can be applied to both vertical and horizontal multi-view data distributions.
% as well as multi-view sequential data.
(3) Deep learning techniques are applied to FedMVL, including but not limited to (a) Deep structured semantic models are used to map users and items to a shared semantic space within a federated multi-view setting \cite{huang2020federated}; (b) DeepMood architecture is used for late fusion in a federated learning setting at the session level \cite{xu2021fedmood}.

Although existing FedMVL methods have designed appropriate frameworks according to the different distributions or characteristics of multi-view data, most studies have focused on labeled data and are not directly applicable to unsupervised multi-view environments. 
In addition, all FedMVL methods assume complete information, but in the real world, not all samples have complete views due to the data storage and usage of multiple clients.
% due to the distributed collection and storage of data.
Unlike previous methods, our method can adapt to unsupervised multi-view environments and address the issue of incomplete views for samples. 
Moreover, we design a mechanism to discover and leverage global self-supervision information, which enhances the quality of the local model for each client and yields a high-quality global clustering structure on the server.

\section{Methodology}

\subsection{Problem Setting}
In this paper, we propose a novel federated deep multi-view clustering method, which can collaborate multi-view data distributed across different clients to mine complementary cluster structures, while addressing data incompleteness and privacy. 
% This method can adapt well to unsupervised multi-view environments and solve the problem of sample missing. 
We focus on the cross-silo federation learning scenario, where all clients participate in each round of communication. 
In a federated multi-view setting, multi-view data with $M$ views, denoted by $\mathbf{X}=\left\{ \mathbf{X}^1,\mathbf{X}^2,...,\mathbf{X}^{M} \right\}$, are distributed among $M$ silos. For client $m$, its data are represented as $\mathbf{X}^m\in \mathbb{R} ^{N_m\times D_m}$, where $D_m$ is the dimensionality of samples in the $m$-th view and $N_m$ is the number of samples in the $m$-th client, $m = 1,...,M$. 
It should be noted that there are differences in the number of samples, sample features and clustering distribution of each client, but there are also some overlapping samples among clients. In this scenario, we clarify two goals:

\textbf{Global goal.} It is expected to obtain a high-quality global clustering structure on the server that is comparable in performance to a model trained on centralized data collected from clients.
% This objective is consistent with the original goal of federated learning. 
% which aligns with the original goal of federated learning.
% It is expected to obtain a high-quality global clustering structure, which is also the original intention of federated learning. 
% Under the global goal, it is expected that the data distributed on different clients can jointly reveal the global common structure shared by multiple views through collaborative training, and its performance can be close to that of a model trained with centralized data collected from clients.

\textbf{Local goal.} It is expected to improve the clustering performance of each client by considering global information, resulting in better performance than the model trained with each client's data independently.
% It is expected to improve the clustering performance of individual clients. Under the local goal, it is expected that the local model, which considers global information, will have better performance than the model trained with each client's data independently.

\subsection{Local Training}
We construct a local model for each client using the same approach and enhance it by considering the global prototypes $\mathbf{C}$ and global pseudo-labels $\mathbf{P}$ obtained from the server. We analyze the local training process for client $m$ as follows.

Deep autoencoder has been widely employed in various unsupervised environments owing to their ability to effectively capture the essential features of the data \cite{feng2014cross,zhang2020deep,lin2021completer}. 
Therefore, we utilize an autoencoder to project the client's data into a low-dimensional space, preserving the privacy of the original data while capturing informative latent features for clustering.
% due to its capacity to effectively capture essential features of the data \cite{feng2014cross,zhang2020deep,lin2021completer}. 
% Therefore, to capture informative latent features for clustering while preserving the privacy of the original data, we utilize an autoencoder to project the client's data into a low-dimensional space. 
The proposed method can be expressed by minimizing the following reconstruction loss:
% This method enables the learning of embedded features denoted as $\mathbf{Z}^m$, by minimizing the reconstruction loss $\mathcal{L} _{r}^{m}$. The formula can be expressed as
\begin{equation} \label{eq1}
 %    \begin{aligned}
	% \mathcal{L} _{r}^{m}&=\left\| \mathbf{X}^m-D_{\theta ^m}\left( \mathbf{Z}^m \right) \right\| _{F}^{2}\\
	% &=\sum_{i=1}^{N_m}{\left\| \mathbf{x}_{i}^{m}-D_{\theta ^m}\left( E_{\phi ^m}\left( \mathbf{x}_{i}^{m} \right) \right) \right\| _{2}^{2}},\\
 %    \end{aligned}
    \mathcal{L} _{r}^{m}=\left\| \mathbf{X}^m-D_{\theta ^m}\left( \mathbf{Z}^m \right) \right\| _{F}^{2}=\sum_{i=1}^{N_m}{\left\| \mathbf{x}_{i}^{m}-D_{\theta ^m}\left( E_{\phi ^m}\left( \mathbf{x}_{i}^{m} \right) \right) \right\| _{2}^{2}},
\end{equation}
where $\mathbf{Z}^m$ denotes the low-dimensional feature embedding of client $m$,
 $E_{\phi ^m}$ and $D_{\theta ^m}$ denote its encoder and decoder networks, respectively.
 The encoder is $E_{\phi ^m}\left( \mathbf{X}^m;\phi ^m \right)\!:\mathbf{X}^m\in \mathbb{R} ^{N_m\times D_m}\longmapsto \mathbf{Z}^m\in \mathbb{R} ^{N_m\times d_m} $ and the decoder is $D_{\theta ^m}\left( \mathbf{Z}^m;\theta ^m \right)\!:\mathbf{Z}^m\in \mathbb{R} ^{N_m\times d_m}\longmapsto \hat{\mathbf{X}}^m\in \mathbb{R} ^{N_m\times D_m}$, where $d_m$ is the dimensionality of embedded features,  $\phi ^m$ and $\theta ^m$ are learnable parameters of autoencoder network.

Inspired by popular single-view deep clustering methods \cite{xie2016unsupervised,guo2017improved}, we use a parameterized mapping $\mathcal{M} _m\left( \mathbf{Z}^m;\mathbf{U}^m \right)\!:\mathbf{Z}^m\in \mathbb{R} ^{N_m\times d_m}\longmapsto \mathbf{Q}^m\in \mathbb{R} ^{N_m\times K}$ to obtain soft cluster assignments $\mathbf{Q}^m$, where $K$ is the number of categories to be clustered. Concretely,
\begin{equation} \label{eq2}
    q_{ij}^{m}=\frac{\left( 1+\left\| \mathbf{z}_{i}^{m}-\mathbf{u}_{j}^{m} \right\| _{2}^{2} \right) ^{-1}}{\sum_{j=1}^K{\left( 1+\left\| \mathbf{z}_{i}^{m}-\mathbf{u}_{j}^{m} \right\| _{2}^{2} \right) ^{-1}}}\in \mathbf{Q}^m,
\end{equation}
where $q_{ij}^{m}$ is the probability that the embedded feature $\mathbf{z}_{i}^{m}$ is assigned to the $j$-th cluster, $\mathbf{U}^m=\left[ \mathbf{u}_{1}^{m};\mathbf{u}_{2}^{m};...;\mathbf{u}_{K}^{m} \right] \in \mathbb{R} ^{K\times d_m}$ represent the learnable parameters, can be initialized with the global prototypes $\mathbf{C}$.

For client $m$, we can convert the global pseudo-labels $\mathbf{P}$ to supervised information $\mathbf{P}^m$ on that client by a mapping $\mathcal{F} _m\left( \mathbf{P} \right)\!:\mathbf{P}\in \mathbb{R} ^{N\times K}\longmapsto \mathbf{P}^m\in \mathbb{R} ^{N_m\times K}$, where $N$ represents the total number of samples on all clients. Furthermore, the clustering loss between the pseudo-labels $\mathbf{P}^m$ and its own cluster assignment distribution $\mathbf{Q}^m$ is optimized:
\begin{equation}
    \mathcal{L} _{c}^{m}=D_{KL}\left( \mathbf{P}^m\parallel \mathbf{Q}^m \right) =\sum_{i=1}^{N_m}{\sum_{j=1}^K{\mathbf{p}_{ij}^{m}}}\log \frac{\mathbf{p}_{ij}^{m}}{\mathbf{q}_{ij}^{m}}.
\end{equation}

So, the total loss of client $m$ consists of two parts:
\begin{equation} \label{eq4}
    \mathcal{L} ^m=\mathcal{L} _{r}^{m}+\gamma \mathcal{L} _{c}^{m},
\end{equation}
where $\gamma$ is a trade-off coefficient between the clustering and reconstruction losses.
 The reconstruction loss $\mathcal{L} _{r}^{m}$ ensures the representation capability of the embedded features to the client's original data. Optimizing the clustering loss $\mathcal{L} _{c}^{m}$   will make the distribution of $\mathbf{Q}^m$ sharper and mine complementary information from other clients by minimizing the $KL$ divergence between $\mathbf{Q}^m$ and $\mathbf{P}^m$.
% the information of other clients by minimizing the $KL$ divergence between $\mathbf{Q}^m$ and $\mathbf{P}^m$.

\subsection{Global Training}
In our framework, to facilitate information flow, each client uploads its embedded features and cluster assignments to the server. 
The server plays a critical role in discovering global self-supervised information, achieving high-quality global clustering, and addressing the challenges of feature
heterogeneity and incomplete information in multi-view data by utilizing sample alignment and data extension techniques.
% The server is used to discover global self-supervised information, achieve high-quality global clustering, and address feature heterogeneity and incomplete information of multi-view data through sample alignment and data extension.
% Server is used to discover global self-supervised information and obtain a high-quality global clustering structure after sample alignment and data extension, thereby alleviating the feature heterogeneity and incomplete information of multi-view data.

After receiving the cluster assignments from each client, the server averages them to obtain the global cluster assignments:
% After receiving the cluster assignments and averages them to get the global cluster assignments:
\begin{equation} \label{eq5}
    \mathbf{Q}=\sum_{m=1}^M{\mathbf{Q}^m\mathbf{A}^m},
\end{equation}
where $\mathbf{Q}\in \mathbb{R} ^{N\times K}$. It is worth noting that the clusters represented by $\mathbf{Q}^m$ in each client do not necessarily correspond to each other. Therefore, we denote $l_{i}^{m}=\operatorname{argmax}_{j} q_{i j}^{m}$, $q_{i j}^{m} \in \mathbf{Q}^m$ and then treat $l^{1}$ as an anchor to modify $l^{m}$ on the remaining clients by minimizing the following matching formula:
\begin{equation} \label{eq6}
    \begin{aligned}
    & \min _{\mathbf{A}^{m}} \mathbf{M}^{m} \mathbf{A}^{m} \\
    \text { s.t. } & \mathbf{A}^m \left( \mathbf{A}^{m} \right) ^T=\mathbf{I}_K,
    \end{aligned}
\end{equation}
where $\mathbf{A}^m$ is a boolean matrix used to adjust the arrangement of $\mathbf{Q}^m$ and $\mathbf{M}^{m} \in \mathbb{R}^{K \times K}$ denotes the cost matrix. $\mathbf{M}^{m}=\max _{i, j} \tilde{m}_{i j}^{m}-\tilde{\mathbf{M}}^{m}$ and $\tilde{m}_{i j}^{m}=\sum_{n=1}^{N} \mathbb{1}\left[l_{n}^{m}=i\right] \mathbb{1}\left[l_{n}^{1}=j\right]$, where $\mathbb{1}[\cdot]$ represents the indicator function. 
The optimization of Eq. (\ref{eq6}) is performed using the Hungarian algorithm \cite{jonker1986improving}.
% Eq. (\ref{eq6}) is optimized with Hungarian algorithm.

After receiving the embedded features uploaded by each client, the server concatenates them to generate the global features:
\begin{equation}
    \mathbf{Z}=\left[ \mathbf{Z}^1,\mathbf{Z}^2,...,\mathbf{Z}^M \right] \in \mathbb{R} ^{N\times \sum_{m=1}^M{d_m}}.
\end{equation}

On the server, we employ an indicator matrix $\mathbf{H}\in \{0,1\}^{N\times M}$, where $h_{im}\in \mathbf{H}$, $h_{im}=1$ denotes client $m$ has data for the $i$-th sample, and otherwise $h_{im}=0$. Moreover, we denote $\mathbf{Z}=\left[ \mathbf{Z}_C;\mathbf{Z}_I \right]$. For each $\mathbf{z}_{i} \in \mathbf{Z}$, if there exists $\sum_{m=1}^M{h_{im}}=M$, then $\mathbf{z}_{i} \in \mathbf{Z}_{C}$; otherwise $\mathbf{z}_{i} \in \mathbf{Z}_{I}$. 

% Based on the information of the samples overlapped for each client, we can obtain the global prototypes $\mathbf{C}$ by the following objective:
By leveraging the overlapping samples across clients, we can obtain the global prototypes $\mathbf{C}$ using the following objective:
\begin{equation} \label{eq8}
    \min_{\mathbf{C}} \left\| \mathbf{Z}_C-\mathbf{C} \right\| _{F}^{2}\\=\min_{\left\{ \mathbf{c}_j \right\} _{j=1}^{K}} \sum_{\mathbf{z}_i\in \mathbf{Z}_C}{\sum_{j=1}^K{\left\| \mathbf{z}_i-\mathbf{c}_j \right\| _{2}^{2}}},\\
\end{equation}
where $\mathbf{C} \in \mathbb{R}^{K \times \sum_{m=1}^{M} d_{m}}$ and $\mathbf{c}_j=\left[ \mathbf{c}_{j}^{1},\mathbf{c}_{j}^{2},...,\mathbf{c}_{j}^{M} \right]$. The global prototypes represent the shared common pattern among samples belonging to the same cluster, obtained by aligning the overlapping client samples.
% The global prototypes represent the global common structure after aligning various client samples, which is the shared common pattern among the samples belonging to the same cluster.

To extract view-specific patterns $\mathbf{W}$ from each client's data, we utilize the following optimization:
\begin{equation} \label{eq9}
    \begin{aligned}
	&\min_{\mathbf{W}} \left\| \mathbf{Z}_C-\mathbf{WQC} \right\| _{F}^{2}\\
	=&\min_{\left\{ \mathbf{W}^m \right\} _{m=1}^{M}} \sum_{\mathbf{z}_i\in \mathbf{Z}_C}{\sum_{m=1}^M{\left\| \mathbf{z}_{i}^{m}-\mathbf{W}^m\mathbf{q}_i\mathbf{C}^m \right\| _{2}^{2}}},\\
    \end{aligned}
\end{equation}
where $\mathbf{q}_i\in \mathbf{Q}$. 
We can leverage information from the global prototypes $\mathbf{C}$, global cluster assignments $\mathbf{Q}$, and view-specific patterns $\mathbf{W}$ to impute the unavailable embedded features $\mathbf{z}_{i}^{m}$. 
% Then we can combine the information from the global prototypes $\mathbf{C}$, global cluster assignments $\mathbf{Q}$ and the view-specific patterns $\mathbf{W}$ to complete the unavailable embedded features $\mathbf{z}_{i}^{m}$.
Specifically, $\mathbf{QC}$ and $\mathbf{W}$ impute the unavailable embedded features from the perspective of sample commonality and view versatility, respectively.
In this case, when $h_{im}=0$ for $\mathbf{z}_{i}^{m}$, the calculation is as follows:
% In this case, for $h_{im}=0$ corresponding to $mathbf{z}_{i}^{m}$ is calculated by
\begin{equation} \label{eq10}        
\mathbf{z}_{i}^{m}=\mathbf{W}^m\mathbf{q}_i\mathbf{C}^m\in \mathbf{Z}_I.
\end{equation}

% Starting from the global common structure, $\mathbf{z}_{i}^{m}$ synthesizes the commonality of samples and the versatility of views to expand the data effectively. 
By starting with the global common structure, $\mathbf{z}_{i}^{m}$ combines the common characteristics of samples with the versatile features of views, resulting in effective data extension.
% By imputing in this way, sharing information and partially overlapping parts of samples among various clients can be utilized, facilitating the mining of more accurate global pseudo-labels $\mathbf{P}$ later.
This imputation method enables the utilization of shared information and partially overlapping parts of samples among various clients, facilitating the mining of more accurate global pseudo-labels $\mathbf{P}$ later.

We concatenate the embedded features uploaded by each client with the features obtained by expanding the data in Eq. (\ref{eq10}) to update global features $\mathbf{Z}=\left[ \mathbf{Z}_C;\mathbf{Z}_I \right]$. 
Then we adopt $K$-means \cite{macqueen1967classification} on the global features to obtain the global clustering structure and calculate the cluster centroids:
\begin{equation} \label{eq11}
    \min _{\mathbf{c}_{1}, \mathbf{c}_{2}, \ldots, \mathbf{c}_{K}} \sum_{i=1}^{N} \sum_{j=1}^{K}\left\|\mathbf{z}_{i}-\mathbf{c}_{j}\right\|^{2}.
\end{equation}

After that, we can use the Student's $t$-distribution to measure the similarity between global features and cluster centroids as follows:
\begin{equation} \label{eq12}
  s_{ij}=\frac{\left(1+\left\|\mathbf{z}_{i}-\mathbf{c}_{j}\right\|^{2}\right)^{-1}}{\sum_{j}\left(1+\left\|\mathbf{z}_{i}-\mathbf{c}_{j}\right\|^{2}\right)^{-1}} \in \mathbf{S}.
\end{equation}
In this way, the confidence $s_{i j}$ is high when $\mathbf{z}_{i}$ is closer to $\mathbf{c}_{j}$. We use the function $\mathcal{E}(\mathbf{S})$ to enhance the confidence and obtain the global pseudo-labels $\mathbf{P}$:
\begin{equation} \label{eq13}
    p_{i j}=\mathcal{E}\left(\mathbf{s}_{i}\right)=\frac{\left(s_{i j} / \sum_{j} s_{i j}\right)^{2}}{\sum_{j}\left(s_{i j} / \sum_{j} s_{i j}\right)^{2}} \in \mathbf{P}.
\end{equation}

Furthermore, the global clustering predictions are calculated by
\begin{equation} \label{eq14}
    y_{i}=\arg \max _{j}\left(p_{i j}\right).
\end{equation}
In summary, the server effectively utilizes the information uploaded by clients to mine global prototypes and global pseudo-labels based on sample commonality and view versatility, and discovers a clear global clustering structure.

\renewcommand{\algorithmicrequire}{\textbf{Input:}}  % Use Input in the format of Algorithm
\renewcommand{\algorithmicensure}{\textbf{Output:}} % Use Output in the format of Algorithm
\begin{algorithm}[!t]
  \caption{Federated Deep Multi-View Clustering (FedDMVC)} 
  \label{alg1}
  \begin{algorithmic}[1]
    \Require
      Data with $M$ views $\mathbf{X}=\left\{ \mathbf{X}^1,\mathbf{X}^2,... ,\mathbf{X}^{M} \right\}$, which are distributed on $M$ local silos, number of clusters $K$, Epoch $E$.
    \Ensure
       Global clustering predictions.
       \While{not reaching $E$ epochs}
       % client
        \For{$m=1$ to $M$} \textbf{in parallel}
            \If{$E==1$}
                \State Get $\theta ^m,\phi ^m,$ and $\mathbf{U}^m$ by pretraining autoencoder.
            \Else
                \State Update $\mathbf{U}^m$ by global prototypes $\mathbf{C}$.
                \While{not reach the maximum iterations $T_1$}
                    \State Optimize the total loss function by Eq. (\ref{eq4}).
                \EndWhile
            \EndIf
            \State Upload $\mathbf{Z}^m$ and $\mathbf{Q}^m$ to the server.
        \EndFor
        % server
        \State Update global cluster assignments $\mathbf{Q}$ by Eqs. (\ref{eq5})-(\ref{eq6}).
        \State Obtain global prototypes $\mathbf{C}$ by Eq. (\ref{eq8}).
        \While{not reach the maximum iterations $T_2$}
            \State Impute the unavailable embedded features by Eq. (\ref{eq9}).
        \EndWhile
        \State Update global features $\mathbf{Z}$ by Eq. (\ref{eq10}).
        \State Obtain global pseudo-labels $\mathbf{P}$ by Eqs. (\ref{eq11})-(\ref{eq13}).
        \State Distribute $\mathbf{C}$ and $\mathbf{P}$ to each client.
       \EndWhile
       \State Calculate the clustering predictions by Eq. (\ref{eq14}).
  \end{algorithmic}
\end{algorithm}

\subsection{Optimization}
% The detailed optimization procedure is described in Algorithm \ref{alg1}. It mainly contains clients and server two parts. Clients are responsible for parallel training of the local model. 
Algorithm \ref{alg1} provides a detailed description of the optimization procedure, which comprises two main parts: the clients and the server. The clients are responsible for parallel training of the local model.
In the first round, they perform pretraining of the autoencoder. In the following rounds, they use the global self-supervision information discovered by the server to enhance the quality of their local models.
% Clients are responsible for training the local model in parallel. They first pretrain the autoencoder in the first round. In subsequent rounds, they utilize the global self-supervision information discovered by the server to improve the quality of their local models. 
The server aligns and imputes the unavailable embedded features, utilizing the information uploaded by clients to address the issue of incomplete sample overlap. 
% extends the data based on sample commonality and view versatility, using the embedded features and cluster assignments uploaded by clients to solve the problem of incomplete sample overlap. 
In addition, the server discovers the global prototypes and global pseudo-labels from the global features, and obtains global clustering predictions. 
Clients and the server alternately iterate through $E$ epochs.
% , we let $E=2$ for all experiments.

\paragraph{\textbf{Complexity Analysis.}} 
% \textbf{Complexity analysis.}
Suppose $K$, $M$, and $N$ represent the number of clusters, clients and total samples, respectively. Let $H$ denote the maximum number of neurons in autoencoders’ hidden layers, $W$ denote the maximum number of hidden neurons in the network on the server, and $Z$ denote the maximum dimensionality of embedded features. 
Generally $N \gg V, K, M$ holds. In Algorithm \ref{alg1}, for client $m$, the complexities to optimize Eq. (\ref{eq1}) and Eq. (\ref{eq2}) are $O(N H^2)$ and $O(N Z K)$, respectively. 
For server, the complexities to optimize Eq. (\ref{eq6}) and Eq. (\ref{eq9}) are $O(MK^3+NMK)$ and $O(N W)$, respectively, while the complexity to optimize Eq. (\ref{eq11}) is $O(N M Z K)$. 
In conclusion, 
% due to $M$ clients running in parallel, 
the total complexity of our algorithm is $O(N H^2+N MZ K+M K^3+N W)$ in each iteration, which is linear to the data size $N$.

\section{Experiment}
\subsection{Experimental Settings}
\paragraph{\textbf{Datasets.}}
% \textbf{Datasets.}
% As shown in Table \ref{datasets}, 
Our experiments are carried out on four widely used datasets. 
Specifically, \textbf{Reuters} \cite{amini2009learning} contains 1200 articles in 6 categories, with each article written in five different languages and treated as five separate text views.
\textbf{Scene} \cite{fei2005bayesian} includes 4,485 scene images in 15 classes with three views. \textbf{Handwritten Numerals (HW)}\footnote{https://archive.ics.uci.edu/ml/datasets.php} contains 2000 samples in 10 categories corresponding to numerals 0-9, each constituted by the six visual views. \textbf{Fashion-MV} \cite{xiao2017fashion} contains images from 10 categories, where we treat six different styles of one object as six views, to better simulate the federated learning environment with six clients.

% \begin{table}[!h]
% \caption{The statistics of experimental datasets.}
% \label{datasets}
% \begin{tabular}{|c|c|c|c|}
% \hline
% Dataset    & Sample & View & Dimension                          \\ \hline
% Reuters    & 1200   & 5    & {[}2000, 2000, 2000, 2000, 2000{]} \\ \hline
% Scene      & 4485   & 3    & {[}20, 59, 40{]}                     \\ \hline
% HW         & 2000   & 6    & {[}216, 76, 64, 6, 240, 47{]}           \\ \hline
% Fashion-MV & 10000  & 6    & {[}784, 784, 784, 784, 784, 784{]}      \\ \hline
% \end{tabular}
% \end{table}

Note that in our federated setting, multiple views of these datasets are distributed among different clients and are isolated from each other.
In addition, to evaluate the effectiveness of our method in handling incomplete multi-view data, we randomly remove some samples from arbitrary views, resulting in the incomplete dataset, following \cite{yang2022robust}. 
Also, we define the sample overlapping rate  $\delta=m / n$ among clients, where $n$ is the size of the dataset and $m$ is the number of samples with fully overlapping views for all clients.

\paragraph{\textbf{Comparing Methods.}}
% \textbf{Comparing Methods.}
% We choose some relevant algorithms as comparison methods. 
We select several pertinent algorithms to serve as comparison methods.
Since our method is essentially distributed, we include two distributed multi-view clustering methods as comparison methods, i.e., RMKMC \cite{cai2013multi} and CaMVC \cite{huang2018robust}. 
% i.e., RMKMC (robust multi-view $k$-means clustering \cite{cai2013multi}) and CaMVC (robust multi-view clustering with capped-norm \cite{huang2018robust}). 
Likewise, our method can be applied to IMVC for handling incomplete multi-view data. We compare our method with five state-of-the-art IMVC methods,
i.e., CDIMC-net \cite{wen2020CDIMC}, GIMC-FLSD \cite{wen2020generalized}, HCP-IMSC \cite{li2022high}, IMVC-CBG \cite{wang2022highly} and DSIMVC \cite{tang2022deep}.
% i.e., CDIMC-net (cognitive deep incomplete multi-view clustering network \cite{wen2020CDIMC}), GIMC-FLSD (generalized incomplete multiview clustering with flexible locality structure diffusion \cite{wen2020generalized}), HCP-IMSC (high-order correlation preserved incomplete multi-view subspace clustering \cite{li2022high}), IMVC-CBG (highly-efficient incomplete large-scale multi-view clustering with consensus bipartite graph \cite{wang2022highly}) and DSIMVC (deep safe incomplete multi-view clustering \cite{tang2022deep}).

For fair comparisons, we conduct FedDMVC and baselines under two settings, i.e., $\delta=0.5$ (denoted by Partially) and $\delta=1$ (denoted by Fully). 
As the first two baselines are unable to handle partially overlapping data directly, we preprocess them by filling the incomplete parts with the mean value of the entire view.
% Since the first two baselines cannot directly handle partially overlapping data, we preprocess them by filling the incomplete parts with the mean value of the entire view.

% \paragraph{\textbf{Implementation Details}}
% % \textbf{Implementation Details.}
% Our methods are implemented using PyTorch and the Flower federated learning framework \cite{beutel2020flower}. We used the same fully connected (Fc) autoencoder structure for all four datasets, following \cite{guo2017improved}. The encoder structure for each client is $\text { Input- } \mathrm{Fc}_{500}-\mathrm{Fc}_{500}-\mathrm{Fc}_{2000}-\mathrm{Fc}_{10}$, and the decoder is symmetric with the encoder. 
% All the autoencoders are pretrained for 500 epochs and the dimensionality of all clients’ embedded features is reduced to 10. Also, we set trade-off coefficient $\gamma=0.1$, use the batch size of 256, set the number of local epochs to 300 and communication rounds $E$ to 2. 
% For a fair comparison, all baselines are tuned to the best performance according to the corresponding papers.

\paragraph{\textbf{Evaluation Metrics.}}
% \textbf{Evaluation Metrics}
We evaluate the clustering effectiveness using three metrics: clustering accuracy (ACC), normalized mutual information (NMI), and adjusted rand index (ARI). A higher value for each metric indicates better clustering performance.

% \begin{figure}[!h]
%   \centering
%   \subfigure[Server]{\includegraphics[width=0.35\textwidth,height=0.1\textheight]{fig/server.png}}
%   \subfigure[Client 1]{\includegraphics[width=0.15\textwidth]{fig/c1.png}}
%   \subfigure[Client 2]{\includegraphics[width=0.15\textwidth]{fig/c2.png}}
%   \subfigure[Client 3]{\includegraphics[width=0.15\textwidth]{fig/c3.png}}\\
%   \subfigure[Client 4]{\includegraphics[width=0.15\textwidth]{fig/c4.png}}
%   \subfigure[Client 5]{\includegraphics[width=0.15\textwidth]{fig/c5.png}}
%   \subfigure[Client 6]{\includegraphics[width=0.15\textwidth]{fig/c6.png}}\\
%   \caption{Visualization of the clustering results on HW with the overlapping rate of 0.5.}
%   \label{visual}
% \end{figure}

\begin{table*}[!t]
\caption{Experiments on four datasets. The best result in each column is shown in bold and the second-best is underlined.}
\label{results}
\begin{tabular}{@{}llcccccccccccc@{}}
\toprule
\multicolumn{1}{l}{}                              &                           & \multicolumn{3}{c}{Reuters}                                                                   & \multicolumn{3}{c}{Scene} & \multicolumn{3}{c}{HW}   & \multicolumn{3}{c}{Fashion-MV} \\ \cmidrule(l){3-14} 
\multicolumn{1}{l}{\multirow{-2}{*}{Overlapping}} & \multirow{-2}{*}{\quad Methods} & ACC                           & NMI                           & ARI                           & ACC     & NMI    & ARI    & ACC    & NMI    & ARI    & ACC      & NMI      & ARI      \\ \midrule
                                                  & RMKMC \cite{cai2013multi}            & 0.324                        & 0.178                        & 0.054                        & 0.276  & 0.252 & 0.118 & 0.648 & 0.628 & 0.508 & 0.550 & 0.654 & 0.467         \\
                                                  & CaMVC \cite{huang2018robust}                & 0.313                        & 0.171                        & 0.056                        & 0.296  & 0.293 & 0.147 & 0.730 & 0.673 & 0.585 &  0.501 & 0.582 & 0.391         \\ \cmidrule(l){2-14}
                                                  & CDIMC-net \cite{wen2020CDIMC}        & 0.179                        & 0.040                        & 0.001                        & 0.306  & \uline{0.319} & \uline{0.153} & 0.798 & \uline{0.820} & \uline{0.736} & 0.604   & 0.701   & 0.522   \\
                                                  & GIMC-FLSD \cite{wen2020generalized}             & \uline{0.473}                        & \uline{0.274}                        & \uline{0.202}                        & 0.300  & 0.264 & 0.135 & 0.242 & 0.163 & 0.033 &  0.709 &  0.738 & 0.603        \\
                                                  & HCP-IMSC \cite{li2022high}           & 0.438                        & 0.261                        & 0.178                        & \uline{0.325}  & 0.273 & 0.143 & \uline{0.809} & 0.778 & 0.719 & --          & --         & --         \\
                                                  & IMVC-CBG \cite{wang2022highly}          & 0.364                        & 0.213                        & 0.088                        & 0.268        & 0.270       & 0.144       & 0.471       & 0.473       & 0.237       & 0.468         & 0.439         & 0.202         \\
                                                  & DSIMVC \cite{tang2022deep}            & 0.421                        & 0.256                        & 0.187                        & 0.278  & 0.304 & 0.145 & 0.762 & 0.736 & 0.650 & \uline{0.800}   & \textbf{0.801}  & \uline{0.665}   \\
\multirow{-8}{*}{\quad Partially}                       & FedDMVC (ours)                      & \textbf{0.566} & \textbf{0.299} & \textbf{0.249} & \textbf{0.393}  & \textbf{0.343} & \textbf{0.225} & \textbf{0.893} & \textbf{0.824} & \textbf{0.790} & \textbf{0.820}   & \uline{0.785}   & \textbf{0.690}   \\ \midrule
                                                  & RMKMC \cite{cai2013multi}            & 0.384 & 0.244 & 0.148 & \uline{0.407}  & 0.406 & \uline{0.230} & 0.741 & 0.739 & 0.636 & 0.532         & 0.737   & 0.556         \\
                                                  & CaMVC \cite{huang2018robust}                 & 0.395                        & 0.261                        & 0.166                        & 0.370  & 0.368 & 0.203 & 0.769 & 0.766 & 0.684 &  0.500  &  0.687 & 0.510         \\ \cmidrule(l){2-14}
                                                  & CDIMC-net \cite{wen2020CDIMC}        & 0.356                        & 0.164                        & 0.092                        & 0.387  & \uline{0.407} & 0.193 & \uline{0.845} & \uline{0.901} & \uline{0.826} & 0.696   & 0.801   & 0.642   \\
                                                  & GIMC-FLSD \cite{wen2020generalized}             & 0.475                        & 0.287                        & \uline{0.205}                        & 0.347  & 0.370 & 0.186 & 0.422 & 0.474 & 0.298 &  0.787 & 0.827 & 0.729         \\
                                                  & HCP-IMSC \cite{li2022high}           & 0.418                        & 0.251                        & 0.166                        & 0.380  & 0.330 & 0.183 & 0.826 & 0.793 & 0.743 & --         & --         &  --        \\
                                                  & IMVC-CBG \cite{wang2022highly}          & \uline{0.460}                        & \uline{0.289}                        & 0.156                        & 0.300  & 0.316 & 0.164 & 0.604       & 0.618 & 0.480       & 0.585  & 0.594         & 0.426         \\
                                                  & DSIMVC \cite{tang2022deep}            & 0.434                        & 0.272                        & 0.204                        & 0.284  & 0.322 & 0.152 & 0.817 & 0.792 & 0.735 & \uline{0.905}   & \textbf{0.915}   & 0.853   \\
\multirow{-8}{*}{\quad Fully}                           & FedDMVC (ours)                        & \textbf{0.655}                        & \textbf{0.419}                        & \textbf{0.364}                        & \textbf{0.451}  & \textbf{0.429} & \textbf{0.280} &\textbf{ 0.965} & \textbf{0.925} & \textbf{0.924} & \textbf{0.925}   & \uline{0.904}   & \textbf{0.856}   \\ \bottomrule
\end{tabular}

\end{table*}

\begin{figure*}[!h]
    \centering
    \includegraphics[width=18cm]{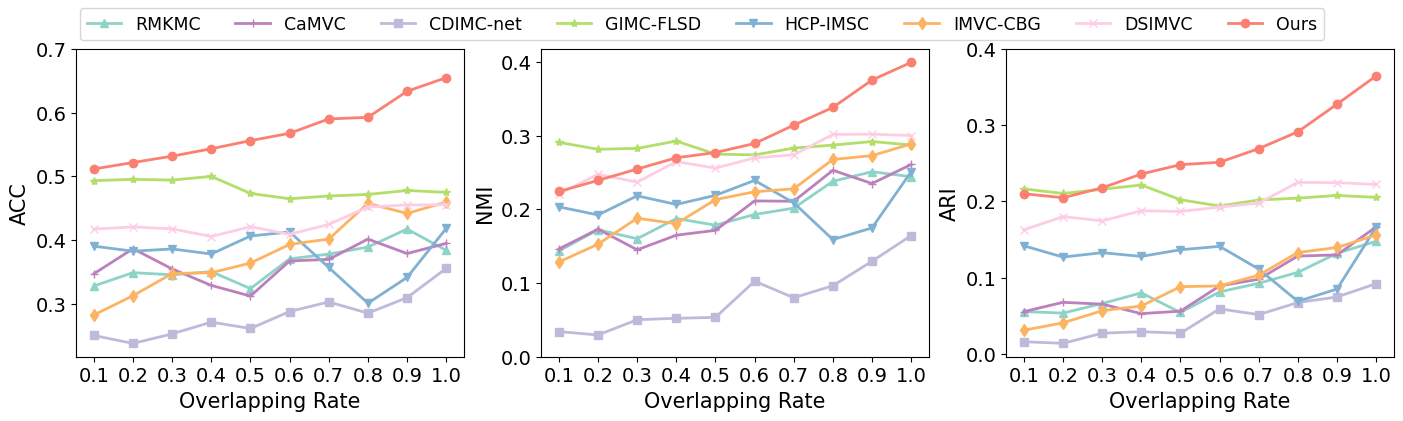}
    \caption{Performance analysis on Reuters with different overlapping rates.}
    \label{fig:rate}
\end{figure*}

\subsection{Clustering Results}
% \paragraph{\textbf{Comparison with Baselines.}}
% \textbf{Comparison with Baselines.}
Table \ref{results} shows the quantitative comparison of FedDMVC and baseline models in the Partially and Fully scenarios. 
% (with the overlapping rate of 50\%)
% (with the overlapping rate of 100\%)
% where the best results are indicated in bold in each column and the second best results are underlined. 
% Due to the high algorithmic complexity, HCP-IMSC could not be executed on the Fashion-MV.
Due to the high algorithmic complexity, HCP-IMSC was unable to be executed on the Fashion-MV dataset.
From Table \ref{results}, we can observe that the proposed method outperforms all baseline models for different scenarios on all datasets.
Compared with the second-best methods CDIMC-net, GIMC-FLSD and DSIMVC, FedDMVC has considerable improvements especially on Reuters, Scene and HW. 
The results demonstrate that our method is effective in handling both complete and incomplete information, while ensuring data privacy in a federated setting.
Particularly in handling incomplete information, the superior performance of FedDMVC validates the effectiveness of our proposed strategy of utilizing sample commonality and view versatility for data extension.

% Furthermore, to further investigate the robustness of our proposed method, 
To further investigate the robustness of our proposed method, we conduct experiments on Reuters with overlapping rates varying from 0.1 to 1 with an interval of 0.1.  
As shown in Figure \ref{fig:rate}, our FedDMVC significantly outperforms the baseline methods across all overlapping rates. Moreover, the performance of FedDMVC shows substantial improvement with increasing overlapping rates. 
The results indicate that FedDMVC is robust to varying degrees of sample overlapping across clients. Additionally, FedDMVC can effectively estimate the data distribution by leveraging available information, even when the overlapping rate is low.

% \paragraph{\textbf{Visualization of Clustering Results.}}
% % \textbf{Visualization of Clustering Results.}
% We use t-SNE \cite{van2008visualizing} to visualize the clustering results on HW, as shown in Figure \ref{visual}.
% The colors in the visualization correspond to the labels of the different nodes, and the final clustering structures of the clients and server are clearly visible. 
% It is shown that FedDMVC can discover the complete global clustering structure and clear local clustering structures.

\subsection{Model Analysis}

\begin{figure*}[!h]
    \centering
    \includegraphics[width=18cm]{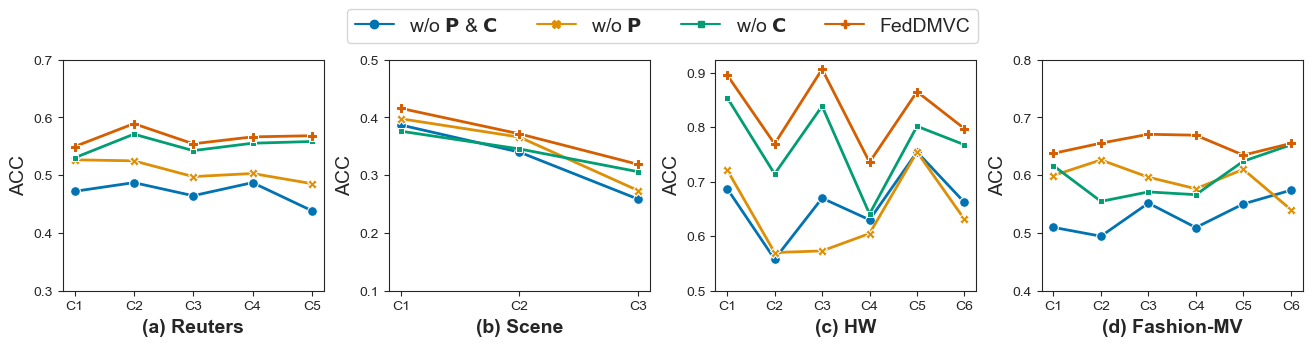}
    \caption{Global information ablation experiments for each client on four datasets with the overlapping rate of 0.5.}
    \label{fig:ablation}
\end{figure*}

\begin{table*}[!t]
\caption{Ablation experiments of the data extension process on the server on four datasets with the overlapping rate of 0.5.}
\label{abla}
\begin{tabular}{c|c|ccc|ccc|ccc|ccc}
\hline
    &             & \multicolumn{3}{c|}{Reuters} & \multicolumn{3}{c|}{Scene} & \multicolumn{3}{c|}{HW}  & \multicolumn{3}{c}{Fashion-MV} \\ \hline
    & Variants    & ACC      & NMI     & ARI     & ACC     & NMI     & ARI    & ACC    & NMI    & ARI    & ACC      & NMI      & ARI      \\ \hline
(A) & w/o $\mathbf{QC}$ \& $\mathbf{W}$ & 0.3375   & 0.1138  & 0.0831  & 0.2415  & 0.1968  & 0.0949 & 0.4910  & 0.3741 & 0.2752 & 0.3322   & 0.3312   & 0.1974   \\ 
(B) & w/o $\mathbf{W}$       & 0.5550    & 0.2945  & 0.2474  & 0.3426  & 0.3046  & 0.1701 & 0.8305 & 0.7979 & 0.7510  & 0.7058    & 0.7623   & 0.6214   \\ 
(C)    & FedDMVC     & \textbf{0.5658}   & \textbf{0.2993}  & \textbf{0.2494}  & \textbf{0.3927}  & \textbf{0.3426}  & \textbf{0.2248} & \textbf{0.8928} & \textbf{0.8071} & \textbf{0.7899} & \textbf{0.8203}   & \textbf{0.7848}   & \textbf{0.6900}   \\ \hline
\end{tabular}
\end{table*}

\paragraph{\textbf{Ablation Study in Each Client's Local Model.}}
% \textbf{Ablation Experiments in Each Client's Local Model}
To further validate the effectiveness of the global information included in our proposed method on the local models of each client, we conduct an ablation study, as shown in Figure \ref{fig:ablation}. 
% In the figure, different colored lines represent different scenarios, where  w/o (=without) denotes that the global information is not included in the method. 
The figure depicts different scenarios, with different colored lines representing each scenario. The label "w/o" denotes the absence of global information in the method.
If the client does not consider any global information, it is equivalent to using autoencoder to extract the embedded features of its own raw data and then performing local clustering. 
If the client does not consider the global pseudo-labels $\mathbf{P}$, it corresponds to only using Eq. (\ref{eq1}) for optimization. 
If the client does not consider the global prototype $\mathbf{C}$, it means that the client uses Eq. (\ref{eq4}) for optimization but still updates the clustering mapping using local centroids. 
The results indicate that incorporating both global pseudo-labels $\mathbf{P}$ and global prototypes $\mathbf{C}$ is advantageous for improving local clustering performance. Furthermore, it is observed that $\mathbf{P}$ has a greater influence on optimizing the local clustering structure than $\mathbf{C}$.
% The results show that the inclusion of both global pseudo labels $\mathbf{P}$ and global prototypes $\mathbf{C}$ is beneficial for local clustering, and that $\mathbf{P}$ has a greater impact on optimizing the local clustering structure relative to $\mathbf{C}$.

\paragraph{\textbf{Variants of Data Extension Process on the Server.}}
To further verify the effectiveness of our proposed method's data extension process on the server, we conduct ablation studies on Eq. (\ref{eq8}) and Eq. (\ref{eq9}). 
% which correspond to mining data sample commonality and view versatility, respectively.
% represent the mining of data sample commonality and view versatility, respectively. 
Table \ref{abla} shows the global clustering results with different variants included. Similarly, w/o represents that the variants are not included in the method. 
(A) represents not using any strategy to impute unavailable embedded features.
% that do not overlap with each other. 
In this case, we directly use the global clustering distribution $\mathbf{Q}$ obtained from Eq. (\ref{eq5}) to obtain global information and global clustering structure. 
(B) considers the commonality of all samples, but lacks estimates of different views from different clients. The results analysis shows that (B) outperforms (A), indicating that $\mathbf{QC}$ estimates sample commonality and is representative of some data features. 
(C) consists of the complete components of our method and outperforms (B). 
% Our method estimates sample commonality while also considering that different views from different clients should exhibit distinct data distribution characteristics.
% The reason is that our method, based on estimating sample commonality, starts from the idea that different views from different clients should have different data distribution characteristics. 
By considering sample commonality and view versatility, we achieve high-quality data extension and obtain a clear global clustering structure.

\begin{figure}[!h]
    \centering
    \includegraphics[width=8.5cm]{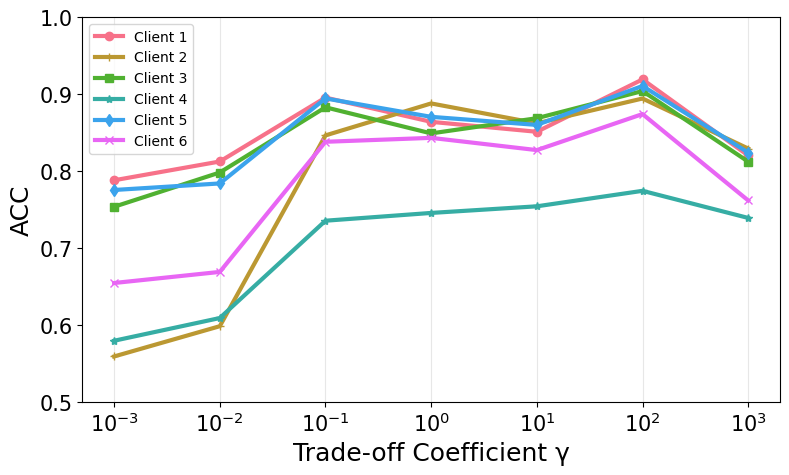}
    \caption{ACC with different $\gamma$ on HW when $\delta=0.5$.}
    \label{fig:para}
\end{figure}

\paragraph{\textbf{Parameter Analysis.}}
% During the training process of each client, the loss function is defined as Eq. (\ref{eq4}), where $\gamma$ is a trade-off coefficient that balances the clustering and reconstruction losses. 
Throughout the training process of each client, the loss function defined in Eq. (\ref{eq4}) incorporates a trade-off coefficient parameter,  $\gamma$, which serves to balance the clustering and reconstruction losses. Here we test the sensitivity of this parameter by varying $\gamma$ from $\left[ 10^{-3},10^{-2},... ,10^3 \right] $.
As shown in Figure \ref{fig:para}, the $\gamma$ range between $\left[10^{-1}, 10^2\right]$ is found to be robust for each client in FedDMVC. 
This indicates that each client needs to consider both losses to achieve a better clustering structure, and highlights the importance of considering global information. 
Without loss of generality, we set $\gamma=0.1$ for all datasets in our experiments.

\begin{figure}[!t]
    \centering
    \includegraphics[width=8.5cm]{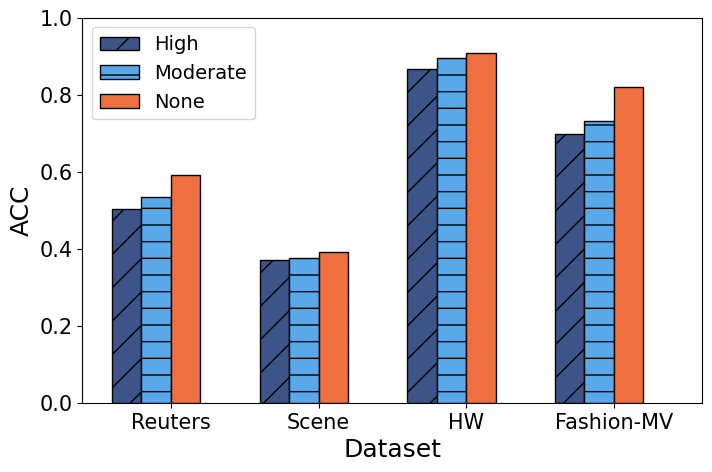}
    \caption{Sensitivity to imbalanced sample sizes among clients on four datasets with the overlapping rate of 0.5.}
    \label{fig:non-iid}
\end{figure}

\paragraph{\textbf{Attributes of Federated Learning.}}
To explore the heterogeneity of sample sizes among clients in federated learning, we introduce Dirichlet distribution when constructing incomplete datasets. 
A smaller Dirichlet parameter $\alpha$ leads to more heterogeneous splits, resulting in highly imbalanced sample sizes among clients. 
Figure \ref{fig:non-iid} illustrates three levels of heterogeneity by setting $\alpha$ to $10^{-2}$ (high), $10^{0}$ (moderate), and $10^{2}$ (none) on four datasets. 
The results show that FedDMVC performs well even in highly heterogeneous scenarios, with only a slight decrease in performance.

\section{Conclusion}
In this paper, we propose a novel federated deep multi-view clustering method, which can collaborate multi-view data stored in different clients to mine complementary cluster structures.
% to obtain a complete and clear global clustering structure. 
% Our proposed method is suitable for scenarios where the $M$ views are distributed across $M$ devices. 
% To address challenges of 1) Views on different clients usually have feature heterogeneity, 
Firstly, we construct global self-supervised information on the server and explore complementary cluster structures across multiple views from multiple clients.
% we construct autoencoders and clustering mappings at each client to mine the embedded features of raw data, while using global self-supervised information to assist local model training on each client. 
% To tackle the challenge of 2) the samples on multiple clients do not completely overlap, 
Furthermore, we propose sample alignment and data extension to impute incomplete data based on sample commonality and view versatility. 
% we perform data extension based on sample commonality and view versatility,  discovering a clear and complete global clustering structure at the server.  
More importantly, the process of discovering and utilizing global self-supervised information enables the flow and sharing of information across clients in a privacy-preserving manner. 
% thus reducing the heterogeneity of multi-view data and utilizing their complementarity. 
Numerous experiments demonstrate that our method outperforms centralized methods that cannot protect data privacy, demonstrating the effectiveness of our proposed method.
% Numerous experiments show that our method outperforms centralized methods which cannot protect data privacy, demonstrating the effectiveness of our proposed method.

\begin{acks}
This work was supported in part by National Key Research and Development Program of China (2020YFC2004300 and 2020YFC2004302), National Natural Science Foundation of China (61971052), Lehigh’s grants (S00010293 and 001250), and National Science Foundation (MRI 2215789 and IIS 1909879).
\end{acks}

%% the bibliography file

% \bibliographystyle{unsrt}
\bibliographystyle{ACM-Reference-Format}
\balance
\bibliography{sample-base}

\appendix

\end{document}